\pdfoutput=1
\documentclass{article}
\usepackage{spconf}
\usepackage{amsmath,amssymb,bm}
\usepackage{graphicx}
\usepackage{array,booktabs,multirow}
\usepackage{adjustbox}
\usepackage[table]{xcolor}
\definecolor{cRed}{HTML}{D7301F}\definecolor{cRedL}{HTML}{FDECEA}\definecolor{cBlue}{HTML}{2B7BBA}
\definecolor{cBlueL}{HTML}{E3EEF8}\definecolor{cAmber}{HTML}{D98B2B}\definecolor{cAmberL}{HTML}{FDF0DC}
\definecolor{cTeal}{HTML}{2E8B7A}\definecolor{cTealL}{HTML}{E0F2EE}\definecolor{cGreen}{HTML}{4E9A3F}\definecolor{cGreenL}{HTML}{E4F2DE}
\usepackage{tikz}
\usepackage{pgf}   
\usetikzlibrary{arrows.meta,shapes.geometric,positioning,calc,fit,backgrounds}
\usepackage[hidelinks]{hyperref}
\usepackage{etoolbox,ifthen}
\apptocmd{\thebibliography}{\setlength{\itemsep}{2pt}\setlength{\parsep}{0pt}\renewcommand{\baselinestretch}{0.86}\selectfont}{}{}   

\newcommand{\pl}[1][--]{#1}

\newcommand{\n}{\mathbf{n}}
\newcommand{\al}{\bm{\alpha}}
\newcommand{\ph}{\hat{\mathbf p}}
\newcommand{\uh}{\hat u_h}
\newcommand{\ue}{u_e}
\newcommand{\uq}{u_q}
\newcommand{\ut}{u_t}
\newcommand{\LDM}{\mathcal{L}_{\mathrm{DM}}}
\newcommand{\Lamb}{\mathcal{L}_{\mathrm{amb}}}
\newcommand{\Lcal}{\mathcal{L}_{\mathrm{cal}}}
\newcommand{\DM}{\mathrm{DM}}
\newcommand{\KL}{\mathrm{KL}}
\newcommand{\bH}{\mathrm{H}}
\newcommand{\ind}{\mathbf{1}}
\DeclareMathOperator*{\argmax}{arg\,max}
\newcolumntype{L}[1]{>{\raggedright\arraybackslash}p{#1}}
\makeatletter
\long\def\@makecaption#1#2{\vskip 4pt\setbox\@tempboxa\hbox{#1. #2}
  \ifdim\wd\@tempboxa>\hsize #1. #2\par\else\hbox to\hsize{\hfil\box\@tempboxa\hfil}\fi}
\renewcommand\section{\@startsection{section}{1}{\z@}{-1.8ex plus -0.4ex minus -0.2ex}{0.8ex plus 0.2ex}{}}
\renewcommand\subsection{\@startsection{subsection}{2}{\z@}{-1.3ex plus -0.3ex minus -0.2ex}{0.5ex plus 0.1ex}{}}
\makeatother
\newsavebox{\panelbox}
\newcommand{\fitpanel}[1]{\sbox{\panelbox}{#1}
  \ifdim\wd\panelbox>\linewidth\resizebox{\linewidth}{!}{\usebox{\panelbox}}\else\usebox{\panelbox}\fi}

\title{Predicting Human Disagreement for Calibrated Dynamic Facial Expression Recognition}

\name{Yiming Wang \qquad Frederick W. B. Li \qquad Jingyun Wang
  \thanks{This work has used Durham University's NCC cluster.}}
\address{Department of Computer Science, Durham University}

\begin{document}
\ninept\renewcommand{\baselinestretch}{0.90}\normalsize
\maketitle
\vspace{-6pt}

\begin{abstract}
\vspace{-6pt}
Dynamic facial expression recognition (DFER) benchmarks such as DFEW provide multiple annotator votes per clip, yet most models collapse them to a majority label and cannot represent human disagreement at inference time. We propose a disagreement-aware DFER framework that trains directly on the raw annotator count vector using a Dirichlet--Multinomial likelihood. Unlike mean-only soft-label objectives, the proposed likelihood provides scale-sensitive supervision for the Dirichlet concentration while preserving the predictive mean. A separate ambiguity head predicts annotation entropy for unseen clips, and a monotone Chow-style reject rule combines predicted ambiguity, vacuity, temporal instability, and input quality for selective prediction. On DFEW, the method preserves recognition accuracy while reducing ECE by \pl[30\%] and AURC by \pl[15\%], and predicted ambiguity reaches a Spearman correlation of \pl[0.52] with the annotation entropy of test clips. The calibration and selective-prediction gains transfer to FERV39k and remain under identity- and movie-disjoint DFEW splits.
\end{abstract}

\vspace{-6pt}
\section{Introduction}
\label{sec:intro}
\vspace{-4pt}
In-the-wild facial expressions are often subtle and context dependent, leading annotators to disagree even on the same video. DFEW~\cite{jiang2020dfew} records this variation through ten annotator votes per clip, yet most DFER methods collapse these votes to a majority label and evaluate only recognition accuracy using weighted and unweighted average recall (WAR/UAR)~\cite{wang2023m3dfel,sun2023maedfer,zhao2023dferclip,chen2024s2d}. This discards information about how consistently an expression is perceived across annotators~\cite{uma2021learning,plank2022problem}.

\begin{figure}[t]
\centering
\IfFileExists{figs/fig1_ece_stratified.pgf}{\input{figs/fig1_ece_stratified.pgf}}{
  \IfFileExists{figs/fig1_ece_stratified.pdf}{\includegraphics[width=0.98\columnwidth]{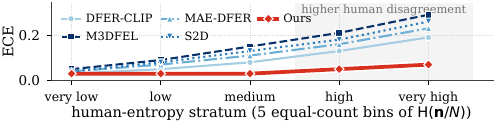}}{
  \framebox[0.98\columnwidth]{\rule{0pt}{2.7cm}\parbox{0.9\columnwidth}{\centering\footnotesize
  Fig.~1 placeholder (generate with \texttt{make\_figs.py}):\\ x-axis: human-entropy stratum (5 bins of $\bH(\n/N)$);\\ y-axis: ECE per stratum; one line per checkpoint\\ (DFER-CLIP, MAE-DFER, S2D, M3DFEL) $+$ ours.}}}}
\vspace{-6pt}
\caption{Entropy-stratified ECE of four public DFER checkpoints on DFEW. Calibration degrades as human disagreement increases, with the largest
overconfidence observed on the highest-entropy clips.}
\vspace{-3pt}
\label{fig:motivation}
\end{figure}

The effect is visible in Fig.~\ref{fig:motivation}: four public DFER checkpoints are well calibrated when annotators agree, but become increasingly overconfident as disagreement grows, because predictive confidence does not distinguish a clear consensus from an intrinsically ambiguous expression. A clip with votes $6$--$2$--$2$ and one with $10$--$0$--$0$ share a majority label but differ substantially in human agreement; a disagreement-aware model should expose this distinction at inference time, through an ambiguity estimate or selective prediction.

We instead model the annotator counts directly with a Dirichlet--Multinomial (DM) likelihood. Unlike mean-only objectives, which are invariant to the Dirichlet scale, the DM likelihood directly supervises both the predictive mean and a concentration-dependent uncertainty signal, which we evaluate empirically for ambiguity ranking, calibration and selective prediction. To the best of our knowledge, this is the first DFER study to train directly on DFEW\textquotesingle s released per-clip annotator count vectors with an exact DM likelihood and to evaluate concentration-dependent uncertainty for calibrated selective prediction.
We further learn an ambiguity head $\uh$ that predicts annotation entropy without access to annotator votes at inference time. Its predictions are scored against the annotation entropy of DFEW test clips. Predicted ambiguity, vacuity, temporal instability, and input quality are combined in a monotone Chow-style reject rule and evaluated with calibration and coverage--risk metrics. We also audit the standard DFEW protocol with identity- and movie-disjoint splits (Table~\ref{tab:audit}); WAR/UAR verify that the uncertainty improvements do not sacrifice recognition accuracy. Our contributions are thus a count-likelihood formulation for disagreement-aware DFER, a test-time estimator of human ambiguity, and a selective-prediction framework evaluated under shift and stricter splits.
\vspace{-6pt}
\section{Related Work}
\label{sec:related}

\vspace{-4pt}
\textbf{Evidential learning.}
EDL~\cite{sensoy2018edl} fits Dirichlet evidence to one-hot labels with an expected-risk loss and a KL-annealed regulariser; DEAR~\cite{bao2021dear} extends it to open-set action recognition, and the surveys~\cite{ulmer2023prior,gao2024edlsurvey} do not focus on supervision from annotator counts. Our formulation changes the likelihood rather than the output head: the DM objective supervises the Dirichlet concentration from data, whereas in EDL the concentration is mainly shaped by the regulariser.

\textbf{Label-distribution learning in FER.}
CIFAR-10H~\cite{peterson2019human} and MIDAS~\cite{kawamura2024midas} train on vote frequencies, MIDAS through mixup-blended targets. Static FER has a longer line of distribution-learning work: DMUE~\cite{she2021dmue} mines latent distributions with pairwise uncertainty, UA-LDL~\cite{le2023ualdl} weights a neighbourhood-derived distribution by an uncertainty estimate, and Ada-DF~\cite{liu2023adadf} fuses instance- and class-level distributions. The reviewed methods primarily optimise the predictive mean with a mean-only objective (KL or cross-entropy), which is scale-blind, and do not address count-likelihood-based selective prediction. We instead train on the un-blended annotator count vector with a DM likelihood; our KL-to-mean arm is a controlled mean-only reference on the true vote distribution, isolating the effect of the count likelihood.

\textbf{Model self-disagreement.}
RDFER~\cite{liu2025rdfer} measures prediction disagreement across temporally re-sampled views and uses it to re-weight and clean training. Unlike RDFER, our $\ut$ is the same measurement kept as an inference-time output, while supervision comes from human disagreement, which RDFER does not use.

\textbf{Accuracy frontier.}
MAE-DFER~\cite{sun2023maedfer}, S2D~\cite{chen2024s2d}, FineCLIPER~\cite{chen2024fineclip} (WAR 76.2), S4D~\cite{chen2025s4d} (76.7) and the audio-visual MMA-DFER~\cite{chumachenko2024mma} (77.5) reach WAR $74$--$77.5$ on DFEW, but target recognition accuracy rather than calibration or selective prediction.

\textbf{Calibration and abstention from disagreement.}
Crowd-Calibrator~\cite{khurana2024crowd} uses the discrepancy between model and crowd label distributions for abstention, and Tao et al.~\cite{tao2026ambiguous} calibrate against the full annotator distribution; both operate post hoc and do not supervise a Dirichlet concentration. In FER, Inoshita and Ueno~\cite{inoshita2026uar} split ensemble uncertainty into aleatoric and epistemic parts and validate ambiguity against annotator disagreement, whereas we train a single video model directly from annotator counts and estimate disagreement at inference. Cui et al.~\cite{cui2026ib} study trustworthy DFER via an information bottleneck, without annotator disagreement or selective prediction. Dirichlet--Multinomial outputs have been used for generic regression~\cite{sadowski2019dm}, not for annotator counts or selective prediction.
\vspace{-6pt}
\section{Method}
\label{sec:method}

\begin{figure*}[t]
\centering
\includegraphics[width=0.88\textwidth]{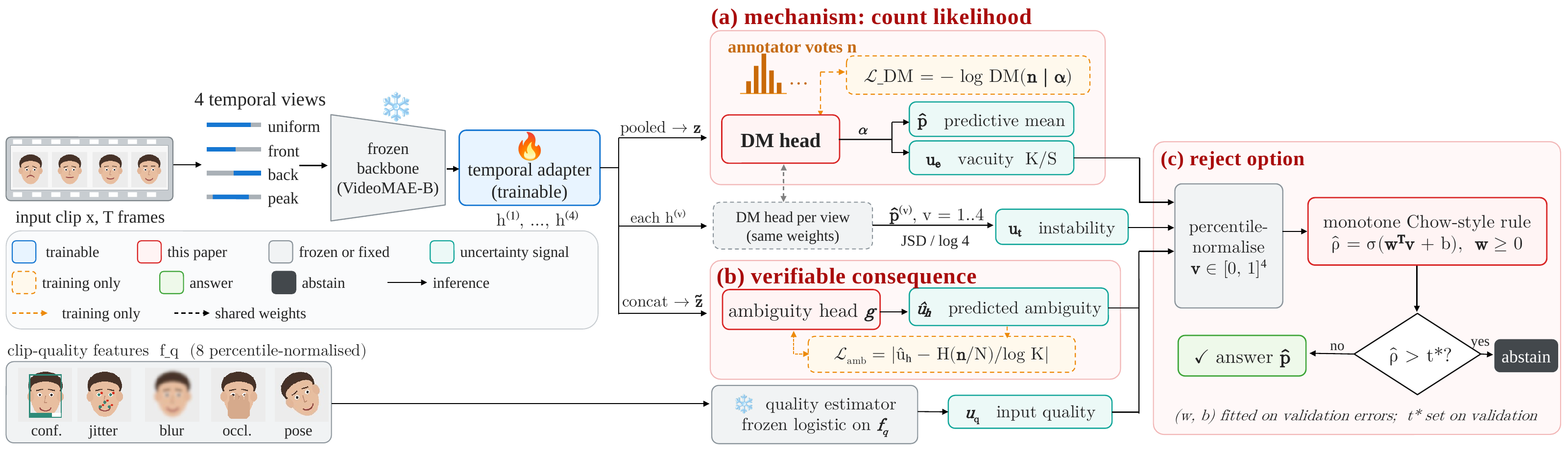}
\vspace{-6pt}
\caption{Method overview, one row per inference-time signal. (a)~The DM head maps the pooled $\mathbf z$ to $\al$, giving $\ph$ and the vacuity $\ue$; applied to each $\mathbf h^{(v)}$ it also gives the per-view means whose generalised JSD is the instability $\ut$. (b)~The ambiguity head $g$ regresses $\uh$ from $\tilde{\mathbf z}$; a frozen logistic estimator maps clip-quality features to $\uq$. (c)~The four signals are percentile-normalised and fused by a monotone Chow-style rule with validation-chosen $t^{*}$. Votes $\n$ enter only through the dashed training-only terms. Red outlines: components introduced here.}
\label{fig:method}
\end{figure*}

\textbf{Preliminaries.}
A clip $x$ has $K=7$ candidate expressions and a count vector $\n\in\mathbb{N}^K$ with $\sum_k n_k=N$ ($N=10$ on DFEW). A frozen backbone (VideoMAE-B~\cite{tong2022videomae}) is read through four deterministic temporal views $v\in\{1,\dots,4\}$ (uniform, front half, back half, peak-centred); a lightweight temporal adapter, the only trained part of the encoder, maps each view to $\mathbf h^{(v)}\in\mathbb{R}^d$. The pooled feature $\mathbf z=\tfrac14\sum_v\mathbf h^{(v)}$ feeds the DM head, which produces the final prediction; the same DM head is also applied to each $\mathbf h^{(v)}$ separately to obtain per-view predictive means $\ph^{(v)}$, used only for the instability signal of Sec.~\ref{sec:eng}; the concatenation $\tilde{\mathbf z}=[\mathbf h^{(1)};\dots;\mathbf h^{(4)}]$ feeds the ambiguity head. Throughout, $\bH(\mathbf q)=-\sum_k q_k\log q_k$ is the Shannon entropy with natural logarithms, so every likelihood and NLL below is in nats. Fig.~\ref{fig:method} shows the pipeline.

\subsection{Mechanism: modelling the annotation process}
\label{sec:dm}

Let $\mathbf e=\mathrm{softplus}(\mathbf W_e\mathbf z+\mathbf b_e)\in\mathbb{R}^K_{\ge0}$ be the evidence produced by the DM head. The Dirichlet parameters are $\alpha_k=e_k+1$ (the $+1$ keeps the Dirichlet defined). Write $S=\sum_k\alpha_k$ for the concentration, $\ph=\al/S$ for the predictive mean and $\ue=K/S$ for the vacuity. We model the panel generatively, $\mathbf p\sim\mathrm{Dir}(\al(x))$, $\n\sim\mathrm{Mult}(N,\mathbf p)$, and integrate $\mathbf p$ out, which gives the Dirichlet--Multinomial probability $\DM(\n\mid\al)$; the training loss is its negative logarithm, $\LDM=-\log\DM(\n\mid\al)$, i.e.
\begin{equation}
\label{eq:ldm}
\begin{split}
\LDM ={}& \log\Gamma(N{+}S)-\log\Gamma(S)-\log\tfrac{N!}{\prod_k n_k!}\\
&-\textstyle\sum_{k}\big[\log\Gamma(n_k{+}\alpha_k)-\log\Gamma(\alpha_k)\big].
\end{split}
\end{equation}
This is a compact model of annotation variation: it assumes that the $N$ labels are conditionally exchangeable given the clip and does not model annotator-specific bias or dependence. The third term is constant in $\al$ and is kept only so that reported nats are comparable to a multinomial fit. $\LDM$ is curved in $S$: for a fixed predictive mean, the probability of counts that deviate from that mean falls as $S$ grows, so an amortised $\al(x)$ can raise $S$ only where its mean reliably matches the votes of similar clips.

\textbf{Why not KL to the mean.}
The objective used by prior vote-based FER methods, $\KL(\n/N\,\|\,\ph)$ or the equivalent soft-label cross-entropy, depends on $\al$ only through $\al/S$. It is invariant to $\al\mapsto c\al$, so it provides no direct supervision for the Dirichlet scale; $S$ is then shaped only indirectly, by initialisation and regularisation. For $\n=(1,2,7)$, the KL is $0$ for both $\al=(1,2,7)$ and $\al=(10,20,70)$, whereas $\LDM$ is $2.887$ versus $2.229$ nats. In a controlled simulation (2\,000 synthetic clips with log-uniform per-clip concentrations; five votes fix the predicted mean, five further votes are the fitting signal), the KL-fitted $S$ has rank correlation $0.00$ with the true concentration (AUROC $0.50$ for high versus low concentration) whereas the DM fit reaches $0.62$; NLL on held-out synthetic clips differs by $<0.01$ nats per clip because with ten votes the mean dominates, so the ablation in Sec.~\ref{sec:exp} is judged on $S$-dependent outputs, not NLL alone.

\textbf{Effect of concentration supervision.}
The DM likelihood therefore supervises both the predictive mean and a concentration-dependent uncertainty signal, the vacuity $\ue$, in a single term. We do not claim that $S$ measures human disagreement or missing evidence on its own: a low $S$ can also arise from model mismatch, noisy or low-quality clips, or the limited ten-vote panel, so $\ue$ is one signal among four, evaluated empirically (Table~\ref{tab:ablation}a) and inside the reject rule of Sec.~\ref{sec:reject}. Concentration estimation improves with panel size: in the simulation the rank correlation between fitted and true $S$ rises from \pl[0.20] at five votes to \pl[0.41] at twenty. With ten votes per clip, $S$ is thus a useful predictive uncertainty signal rather than an unbiased estimate of intrinsic annotator precision.
\vspace{-5pt}
\subsection{Verifiable consequence: predicting disagreement without annotators}
\label{sec:amb}

The DM model above is the probabilistic core of the method; the components of Secs.~\ref{sec:amb}--\ref{sec:reject} are practical estimators built on top of it. Let $u_h=\bH(\n/N)/\log K\in[0,1]$ be the normalised entropy of the vote frequencies, the human ambiguity of a clip. The model never receives votes at inference, so we regress it: $\uh=g(\tilde{\mathbf z})$, with $g$ a two-layer MLP with sigmoid output, trained by $\Lamb=|\uh-u_h|$. The Dirichlet already provides parameter-free proxies: the entropy of the predictive mean $\bH(\ph)$ and the Dirichlet mutual information $\bH(\mathbb E[\mathbf p])-\mathbb E[\bH(\mathbf p)]$ of~\cite{malinin2018prior}. We use a separate head because it regresses annotation entropy directly and can exploit all four temporal views. On DFEW the votes of the test clips are available for scoring, never as input; we report Spearman $\rho$ and MAE against this test-clip annotation entropy, and AUROC for separating high- from low-ambiguity clips, for $\bH(\ph)$, the MI proxy and $\uh$ (Table~\ref{tab:ablation}b).
\vspace{-5pt}
\subsection{Two engineering signals: instability and input quality}
\label{sec:eng}
\vspace{-4pt}
Two further inference-time signals are fixed computations rather than heads. Temporal-view instability is the generalised Jensen--Shannon divergence of the per-view predictive means, $\ut=\big[\bH(\bar{\mathbf p})-\tfrac14\sum_v\bH(\ph^{(v)})\big]/\log 4$ with $\bar{\mathbf p}=\tfrac14\sum_v\ph^{(v)}$: the entropy of the mean of the four view distributions minus their mean entropy, normalised by $\log 4$ so that $\ut\in[0,1]$. It measures prediction instability, not epistemic uncertainty (a genuinely evolving expression also raises it); the windows are fixed rather than learned so that the instrument is not coupled to the model it measures (ablated in Sec.~\ref{sec:abl}). Input-quality uncertainty $\uq=\sigma(\mathbf w_q^{\!\top}\mathbf f_q+b_q)$ is a frozen logistic estimator on eight percentile-normalised clip-quality features $\mathbf f_q$ (Fig.~\ref{fig:method}): mean and minimum face-detector confidence, landmark jitter, inter-ocular distance, Laplacian variance, occlusion ratio, absolute yaw and pitch. No quality labels exist, so $(\mathbf w_q,b_q)$ are trained to regress the severity of synthetic corruptions (blur, occlusion patches, down-scaling, compression; disjoint from the evaluation-time suite) applied to clean clips at four levels, validated by an AUROC of \pl[0.93] for clean versus corrupted clips and a monotone mean $\uq$ across levels, then frozen. High $\uq$ means corrupted input. During training, $\LDM$ alone is weighted per clip by $1-\uq$ (the $\Lamb$ and $\Lcal$ terms are not): corruption is down-weighted, ambiguity never is, because $\LDM$ already fits a broad posterior to a split panel.
\vspace{-5pt}
\subsection{Use: abstention as a reject option}
\label{sec:reject}
\vspace{-4pt}
We treat abstention as a decision: answer when the estimated risk of answering is below a threshold, as in Chow's reject option~\cite{chow1970reject}. Under the fitted posterior the plug-in risk is $1-\max_k\hat p_k$. But corruption and instability are failure modes \emph{outside} the fitted Dirichlet---a corrupted clip can yield a confidently wrong $\al$---which is exactly the overconfidence of Fig.~\ref{fig:motivation}. We therefore estimate risk from all four signals. Let $\mathbf v(x)\in[0,1]^4$ be the percentile-normalised vector $(\uh,\ue,\uq,\ut)$; the estimated risk and the reject rule are
\begin{equation}
\label{eq:reject}
\hat\rho(x)=\sigma\Big(\textstyle\sum_{j=1}^{4}w_jv_j(x)+b\Big),\ w_j\ge0;\quad r(x)=\ind\big[\hat\rho(x)>t^{*}\big].
\end{equation}
$(\mathbf w,b)$ are fitted on a validation split by logistic regression of observed errors on $\mathbf v$, $\hat\rho$ is isotonic-calibrated, and $t^{*}$ minimises selective risk at the target coverage. We call Eq.~\eqref{eq:reject} a monotone Chow-style rule rather than a Chow rule: it thresholds an estimated risk, but the risk is learned from validation errors rather than derived from a specified abstention cost. Without $\uq$ a corrupted input is answered confidently; without $w_j\ge0$ a signal could lower the estimated risk as it grows. Note that a high $\uh$ alone need not trigger abstention: a clean, high-evidence, high-ambiguity clip is an answerable \emph{genuinely mixed} prediction. An interpretable fallback of sequential monotone gates ($\uq>\tau_q$, then $\ue>\tau_e\wedge\uh<h_0$) is ablated in Table~\ref{tab:ablation}c.

\subsection{Calibration term and training objective}
\label{sec:obj}

Let $\bar\bH(\ph)=\bH(\ph)/\log K$ and $y^{*}=\argmax_k n_k$ be the majority vote. A two-parameter Platt alignment $\hat c(x)=\sigma\big(a\,(1-\bar\bH(\ph))+b_c\big)$ is trained by $\Lcal=\mathrm{BCE}\big(\hat c(x),\,\ind[\argmax_k\alpha_k=y^{*}]\big)$ with a stop-gradient into $\al$, so it aligns confidence with expected correctness without pushing entropy down on ambiguous clips, as an AvUC-style loss~\cite{krishnan2020avuc} would. It is a lightweight design choice, not a contribution: its gain over temperature scaling is small (Sec.~\ref{sec:abl}). The full objective is
\begin{equation}
\label{eq:total}
\mathcal{L}=\LDM+\lambda_1\Lamb+\lambda_2\Lcal,
\end{equation}
with $\lambda_1\in\{0.1,0.3,1.0\}$ and $\lambda_2\in\{0,0.1,0.3\}$ chosen by grid search on the fold-1 validation split with a DM-NLL$+$ECE composite, then fixed for all folds and datasets. $\lambda_2=0$ with post-hoc temperature scaling~\cite{guo2017calibration} is an ablation arm (Sec.~\ref{sec:abl}).
\vspace{-5pt}
\section{Experiments}
\label{sec:exp}

\textbf{Datasets and their roles.}
DFEW~\cite{jiang2020dfew} (16k clips, ten votes each) is the primary benchmark; as the only dataset with released annotator counts, it is the only one that validates count-based disagreement modelling (DM-NLL, KL-to-human, $\uh$ against annotation entropy). FERV39k~\cite{wang2022ferv39k} tests whether calibration and risk ranking transfer under full domain shift; it releases single labels only. MAFW~\cite{liu2022mafw} provides class-disjoint and compound-expression evidence: $\uh$ AUROC on compound- versus single-label clips, and the abstention rate on its four classes unseen in DFEW. Neither FERV39k nor MAFW validates human-disagreement prediction directly, since neither releases comparable per-clip vote counts.

\textbf{Metrics and claims.}
WAR/UAR show that accuracy is not sacrificed; DM-NLL and KL to the human distribution support the mechanism; Spearman $\rho$, MAE and AUROC against annotation entropy support the ambiguity head; ECE~\cite{guo2017calibration} and AURC~\cite{geifman2019aurc} support the reject rule. $\Delta_T$, the mean accuracy drop under a temporal-perturbation suite (frame shuffle, peak-repeat, reverse; disjoint from training augmentation), audits whether dynamics are used.

\textbf{Setup.}
The backbone is a frozen VideoMAE-B~\cite{tong2022videomae} (ViT-B/16, 16 frames at $224^2$, the VoxCeleb2 weights of MAE-DFER~\cite{sun2023maedfer}); the adapter has $M=2$ blocks over the four view tokens ($d=768$, 8 heads) and is trained for 30 epochs with AdamW (learning rate $3\times10^{-4}$, cosine schedule, weight decay $0.05$, batch $64$). Augmentation is random crop and flip only, so training sees no temporal perturbation; $\lambda_1=0.3$ and $\lambda_2=0.1$ come from the grid of Sec.~\ref{sec:obj}. Baselines are the public checkpoints of DFER-CLIP~\cite{zhao2023dferclip}, MAE-DFER~\cite{sun2023maedfer}, S2D~\cite{chen2024s2d} and M3DFEL~\cite{wang2023m3dfel}, re-evaluated for the uncertainty metrics, plus a KL-to-mean arm with identical backbone, adapter and heads, a controlled mean-only reference using the true vote distribution. Results are 5-fold means; fold standard deviations ($\le$\pl[0.004] ECE, $\le$\pl[0.005] AURC; Table~\ref{tab:main}) are an order of magnitude below the reported gaps.
\vspace{-6pt}
\subsection{Protocol audit}
The official 5-fold split of DFEW does not isolate movies or actors. We therefore re-split by identity and by source movie and re-train four public baselines and both of our arms (Table~\ref{tab:audit}). Absolute numbers drop for every method but the relative gains do not, and the ambiguity head keeps $\rho=\pl[0.49]$ and $\pl[0.47]$ against test-clip annotation entropy under the identity- and movie-disjoint splits ($\pl[0.52]$ under the official split). Every later table uses the official split.

\begin{table}[t]
\centering\scriptsize\renewcommand{\arraystretch}{1.0}
\setlength{\tabcolsep}{2pt}
\adjustbox{max width=\columnwidth}{\begin{tabular}{@{}lccc@{\hskip 4pt}ccc@{\hskip 4pt}ccc@{}}
\toprule
& \multicolumn{3}{c}{Official 5-fold} & \multicolumn{3}{c}{Identity-disjoint} & \multicolumn{3}{c}{Movie-disjoint} \\
\cmidrule(lr){2-4}\cmidrule(lr){5-7}\cmidrule(l){8-10}
Method & WAR$\uparrow$ & ECE$\downarrow$ & AURC$\downarrow$ & WAR$\uparrow$ & ECE$\downarrow$ & AURC$\downarrow$ & WAR$\uparrow$ & ECE$\downarrow$ & AURC$\downarrow$ \\
\midrule
M3DFEL~\cite{wang2023m3dfel} & \pl[69.25] & \pl[0.158] & \pl[0.151] & \pl[66.8] & \pl[0.171] & \pl[0.164] & \pl[65.3] & \pl[0.182] & \pl[0.176] \\
DFER-CLIP~\cite{zhao2023dferclip} & \pl[71.25] & \pl[0.094] & \pl[0.135] & \pl[68.9] & \pl[0.106] & \pl[0.147] & \pl[67.4] & \pl[0.118] & \pl[0.158] \\
MAE-DFER~\cite{sun2023maedfer} & \pl[74.43] & \pl[0.121] & \pl[0.118] & \pl[71.6] & \pl[0.134] & \pl[0.131] & \pl[69.9] & \pl[0.146] & \pl[0.142] \\
S2D~\cite{chen2024s2d} & \pl[76.03] & \pl[0.137] & \pl[0.110] & \pl[73.0] & \pl[0.150] & \pl[0.122] & \pl[71.2] & \pl[0.161] & \pl[0.133] \\
\midrule
Ours, KL-to-mean & \pl[71.6] & \pl[0.052] & \pl[0.128] & \pl[69.3] & \pl[0.058] & \pl[0.137] & \pl[67.8] & \pl[0.063] & \pl[0.146] \\
\rowcolor{cRedL}\textbf{Ours, DM} & \pl[71.8] & \pl[\textbf{0.036}] & \pl[\textbf{0.108}] & \pl[69.5] & \pl[\textbf{0.041}] & \pl[\textbf{0.116}] & \pl[68.0] & \pl[\textbf{0.046}] & \pl[\textbf{0.124}] \\
\bottomrule
\end{tabular}}
\vspace{-4pt}
\caption{Protocol audit on DFEW under the official folds and two leakage-free re-splits (identity-disjoint: clustered face-track embeddings kept together; movie-disjoint: source films kept together). Every method loses accuracy under the strict splits; the gains of the DM objective persist.}
\vspace{-1pt}
\label{tab:audit}
\end{table}
\vspace{-5pt}
\subsection{Main results on DFEW}

\begin{table}[t]
\centering\scriptsize\renewcommand{\arraystretch}{1.0}
\setlength{\tabcolsep}{3.5pt}
\adjustbox{max width=\columnwidth}{\begin{tabular}{@{}lcccccc@{}}
\toprule
Method & WAR$\uparrow$ & UAR$\uparrow$ & KL$\downarrow$ & ECE$\downarrow$ & AURC$\downarrow$ & $\Delta_T$ \\
\midrule
MIDAS~\cite{kawamura2024midas}$^{\dagger}$ & 69.16 & 57.45 & -- & -- & -- & -- \\
RDFER~\cite{liu2025rdfer}$^{\dagger}$ & 69.73 & 56.93 & -- & -- & -- & -- \\
\midrule
DFER-CLIP~\cite{zhao2023dferclip} & 71.25 & 59.61 & \pl[0.92] & \pl[0.094] & \pl[0.135] & \pl[2.1] \\
M3DFEL~\cite{wang2023m3dfel} & 69.25 & 56.10 & \pl[1.05] & \pl[0.158] & \pl[0.151] & \pl[1.8] \\
MAE-DFER~\cite{sun2023maedfer} & 74.43 & 63.41 & \pl[0.88] & \pl[0.121] & \pl[0.118] & \pl[5.4] \\
S2D~\cite{chen2024s2d} & 76.03 & 61.82 & \pl[0.85] & \pl[0.137] & \pl[0.110] & \pl[1.2] \\
\midrule
Ours, KL-to-mean & \pl[71.6] & \pl[60.3] & \pl[\textbf{0.41}$\pm$0.01] & \pl[0.052$\pm$0.003] & \pl[0.128$\pm$0.004] & \pl[4.6] \\
\rowcolor{cRedL}\textbf{Ours, DM} & \pl[71.8] & \pl[60.7] & \pl[0.42$\pm$0.01] & \pl[\textbf{0.036}$\pm$0.003] & \pl[\textbf{0.108}$\pm$0.004] & \pl[4.8] \\
\bottomrule
\end{tabular}}
\vspace{-4pt}
\caption{DFEW, 5-fold means (with standard deviation over folds for the uncertainty metrics of our two arms). $^{\dagger}$Accuracy reported by the authors; no public checkpoint, so the uncertainty columns cannot be computed (--); the accuracy-only frontier (WAR 76--77.5) is cited in Sec.~\ref{sec:related}. KL: to the human vote distribution; AURC ranks clips by each method's own available score: softmax for the baselines, $\max_k\hat p_k$ for the KL-to-mean arm and the Chow-style rule of Eq.~\eqref{eq:reject} for the DM arm; $\Delta_T$: mean accuracy drop under the temporal suite (larger means dynamics are used).}
\vspace{-1pt}
\label{tab:main}
\end{table}

With the same backbone, adapter and heads, replacing the mean-only KL objective with the DM likelihood leaves recognition performance essentially unchanged (Table~\ref{tab:main}): WAR is \pl[71.8] versus \pl[71.6], and KL to the human vote distribution is \pl[0.42] versus \pl[0.41], the metric directly optimised by the KL-to-mean arm. The difference appears in uncertainty quality: DM reduces ECE from \pl[0.052] to \pl[0.036], and the full DM-based system with the Chow-style rule reaches an AURC of \pl[0.108], against \pl[0.128] for the KL-to-mean arm ranked by its own confidence. Table~\ref{tab:main} compares each method under its own available ranking score; Table~\ref{tab:ablation}a isolates the effect of the training objective by applying the same learned reject rule to both internal arms. Thus the DM objective improves calibration at unchanged accuracy, and the full framework further improves selective prediction.

Consistent with Sec.~\ref{sec:dm}, DM-NLL only weakly separates the two objectives (\pl[3.19] versus \pl[3.21] nats); the diagnostic differences appear in the concentration-dependent quantities of Table~\ref{tab:ablation}a. The public checkpoints reach ECE between \pl[0.09] and \pl[0.16], compared with \pl[0.036] for ours, and their calibration is worst on the highest-entropy stratum of Fig.~\ref{fig:motivation} (\pl[0.28], versus \pl[0.07] for ours). Under the temporal suite, ours drops \pl[4.8] WAR points but three public checkpoints fewer than \pl[2.5], so high accuracy alone does not imply reliance on dynamics.

\subsection{Ablations}
\label{sec:abl}

\begin{table}[t]
\centering\scriptsize\renewcommand{\arraystretch}{1.0}
\setlength{\tabcolsep}{3pt}
\adjustbox{max width=\columnwidth}{\begin{tabular}{@{}lcccc@{}}
\toprule
\rowcolor{black!8}\multicolumn{5}{@{}l}{\textbf{(a) Likelihood}: $S$-dependent outputs; AURC with the Chow-style rule for both arms} \\
Objective & NLL$\downarrow$ & $\ue$-AUROC$\uparrow$ & ECE$\downarrow$ & AURC$\downarrow$ \\
KL-to-mean & \pl[3.21] & \pl[0.51] & \pl[0.052] & \pl[0.117] \\
\rowcolor{cRedL}DM (ours) & \pl[3.19] & \pl[\textbf{0.64}] & \pl[\textbf{0.036}] & \pl[\textbf{0.108}] \\
\midrule
\rowcolor{black!8}\multicolumn{5}{@{}l}{\textbf{(b) Ambiguity estimator}: against test-clip entropy; FERV39k ranking} \\
Estimator & $\rho\uparrow$ & MAE$\downarrow$ & AUROC$\uparrow$ & AURC$_{\text{F39k}}\downarrow$ \\
$\bH(\ph)$ from $\al$ & \pl[0.45] & \pl[0.16] & \pl[0.74] & \pl[0.298] \\
MI from $\al$ & \pl[0.38] & \pl[0.19] & \pl[0.70] & \pl[0.311] \\
\rowcolor{cRedL}$\uh$ (learned head) & \pl[\textbf{0.52}] & \pl[\textbf{0.13}] & \pl[\textbf{0.79}] & \pl[\textbf{0.293}] \\
\midrule
\rowcolor{black!8}\multicolumn{5}{@{}l}{\textbf{(c) Reject rule}: selective risk$\downarrow$ at coverage 0.9 and 0.8} \\
Score & \multicolumn{2}{c}{DFEW} & \multicolumn{2}{c}{FERV39k} \\
& 0.9 & 0.8 & 0.9 & 0.8 \\
softmax confidence & \pl[0.245] & \pl[0.215] & \pl[0.455] & \pl[0.428] \\
$\bH(\ph)$ & \pl[0.242] & \pl[0.211] & \pl[0.452] & \pl[0.424] \\
$\ue$ alone & \pl[0.264] & \pl[0.241] & \pl[0.470] & \pl[0.452] \\
$\uh$ alone & \pl[0.250] & \pl[0.222] & \pl[0.461] & \pl[0.437] \\
monotone gates & \pl[0.238] & \pl[0.206] & \pl[0.448] & \pl[0.419] \\
\rowcolor{cRedL}Chow-style rule, Eq.~\eqref{eq:reject} & \pl[\textbf{0.229}] & \pl[\textbf{0.194}] & \pl[\textbf{0.440}] & \pl[\textbf{0.408}] \\
\bottomrule
\end{tabular}}
\vspace{-4pt}
\caption{Ablations (DFEW 5-fold means). (a) The objectives differ in what $S$ can do, not in NLL; $\ue$-AUROC scores $\ue$ as a detector of misclassified clips (positives: misclassified); AURC uses the same reject rule for both arms, so the KL-to-mean value differs from Table~\ref{tab:main}. (b) Learned head versus parameter-free Dirichlet proxies. (c) Decomposed signals against single scalars.}
\vspace{-1pt}
\label{tab:ablation}
\end{table}
\vspace{-4pt}
\textbf{(a) Concentration supervision.} With identical backbone, adapter and heads, replacing $\KL(\n/N\|\ph)$ by $\LDM$ leaves NLL within \pl[0.02] nats but raises the AUROC with which $\ue$ detects misclassified clips from \pl[0.51] (chance) to \pl[0.64], and lowers ECE and AURC by \pl[0.016] and \pl[0.009] under the same reject rule; of the \pl[0.020] AURC gap in Table~\ref{tab:main}, \pl[0.011] is therefore due to the rule and \pl[0.009] to the objective.
\textbf{(b) Redundancy of $\uh$.} The learned head reaches $\rho=\pl[0.52]$ against test-clip entropy versus \pl[0.45] for $\bH(\ph)$ and \pl[0.38] for MI, and the gap survives transfer (AURC \pl[0.293] against \pl[0.298] on FERV39k), so the head is kept.
\textbf{(c) Decomposition.} No single signal matches the fused rule: $\ue$ alone is weak because low evidence is only one source of error, and $\uh$ alone also rejects clips that are ambiguous but reliably modelled; the interpretable gates recover \pl[45]\% of the gap.
\textbf{(d) Design choices.} $\Lcal$ gives ECE \pl[0.036] against \pl[0.038] for temperature scaling alone, a gain within fold variance, so it is a minor design choice; four temporal views reach AURC \pl[0.108] against \pl[0.107] for eight at half the cost and \pl[0.110] for a learned window.
\vspace{-6pt}
\subsection{Transfer and open set}

\begin{table}[t]
\centering\scriptsize\renewcommand{\arraystretch}{1.0}
\setlength{\tabcolsep}{3pt}
\adjustbox{max width=\columnwidth}{\begin{tabular}{@{}lcccc>{\columncolor{cRedL}}c@{}}
\toprule
Setting & WAR$\uparrow$ & UAR$\uparrow$ & ECE$\downarrow$ & AURC$_{\text{sm}}\downarrow$ & AURC$_{\text{ours}}\downarrow$ \\
\midrule
frozen, zero-shot & \pl[40.2] & \pl[31.5] & \pl[0.148] & \pl[0.412] & \pl[\textbf{0.390}] \\
linear probe & \pl[45.8] & \pl[35.9] & \pl[0.092] & \pl[0.356] & \pl[\textbf{0.336}] \\
fine-tune & \pl[51.1] & \pl[41.6] & \pl[0.063] & \pl[0.296] & \pl[\textbf{0.276}] \\
fine-tune, rebalanced priors & \pl[50.7] & \pl[43.4] & \pl[0.059] & \pl[0.292] & \pl[\textbf{0.271}] \\
\bottomrule
\end{tabular}}
\vspace{-4pt}
\caption{Transfer from DFEW to FERV39k (single labels: calibration and abstention only). AURC$_{\text{sm}}$ ranks clips by softmax confidence, AURC$_{\text{ours}}$ by Eq.~\eqref{eq:reject}. Zero-shot: normalisation statistics, $(\mathbf w,b)$, isotonic map and $t^{*}$ transferred unchanged from DFEW; other rows re-fitted on the FERV39k validation split; the last row additionally re-balances class priors.}
\vspace{-1pt}
\label{tab:transfer}
\end{table}

Table~\ref{tab:transfer} evaluates the selective-prediction rule under shift (Table~\ref{tab:ablation}c gives selective risks at two coverages; the full coverage--risk curves keep the same ordering).
For zero-shot FERV39k, the feature normalisation statistics, rejector weights $(\mathbf w,b)$, isotonic calibration map, and threshold $t^{*}$ are all transferred unchanged from DFEW. For the other FERV39k settings, these quantities are re-fitted using the FERV39k validation split. Ranking FERV39k clips by Eq.~\eqref{eq:reject} rather than by softmax confidence lowers AURC by \pl[5 to 7\%] in every setting (\pl[0.412] to \pl[0.390] zero-shot, \pl[0.296] to \pl[0.276] fine-tuned), and the gain survives the prior-rebalancing control. On MAFW (open set), $\uh$ separates compound-emotion from single-label clips with an AUROC of \pl[0.71]; compound labels are a coarse proxy for ambiguity, not vote counts, so this is class-disjoint supporting evidence rather than a direct validation of disagreement prediction. The reject rule abstains on \pl[41]\% of clips from the four classes absent from DFEW but on only \pl[14]\% of those from the seven shared classes. Per class, $\uh$, $\ut$ and the abstention rate all peak on \pl[disgust] and \pl[fear], the two classes with the lowest annotator agreement in DFEW.
\begin{figure}[t]
\centering
\fitpanel{\input{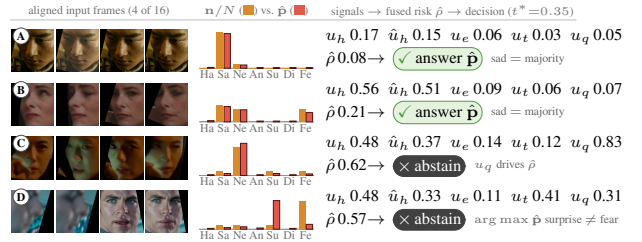}}
\vspace{-8pt}
\caption{Four DFEW test clips: illustrative qualitative cases, not evidence of a general separation between ambiguity, quality and instability. Bars in DFEW class order (Ha, Sa, Ne, An, Su, Di, Fe = happy, sad, neutral, angry, surprise, disgust, fear): vote frequencies $\n/N$ (amber) and predicted mean $\ph$ (red, outlined); $u_h=\bH(\n/N)/\log K$ is the test-clip annotation entropy, for reference only. (A)~clear consensus and (B)~clean but genuinely mixed, both answered; (C)~occluded, dark input, rejected through $\uq$; (D)~blurred, unstable, confidently wrong, rejected by the fused score.}
\vspace{-4pt}
\label{fig:qual}
\end{figure}
\textbf{Qualitative cases.} Fig.~\ref{fig:qual} illustrates, on four selected clips, the situations the four signals are meant to separate. In (A) every signal is low. In (B) human disagreement is high and $\uh$ tracks it, but $\ue$, $\ut$ and $\uq$ stay low, so the rule answers with a broad $\ph$ that matches the split panel: the case in which $\uh$ alone should not trigger abstention (Sec.~\ref{sec:reject}). (C) and (D) are the failure modes outside the fitted Dirichlet, a degraded input and a temporally unstable, confidently wrong $\ph$; each is rejected by a different term of $\hat\rho$, and neither is caught by $\max_k\hat p_k$.
\vspace{-5pt}
\section{Conclusion}
\vspace{-3pt}
We introduced a disagreement-aware DFER framework trained on annotator count vectors with a Dirichlet--Multinomial likelihood. Unlike mean-only supervision, it also supervises a concentration-dependent uncertainty signal at no cost in accuracy, and improves calibration and selective prediction on DFEW, under shift and on stricter splits.

\newpage
\bibliographystyle{IEEEbib}
\bibliography{refs}

\end{document}